\documentclass[11pt,a4paper]{article}
\usepackage{authblk}
\usepackage[hyperref]{acl2021}
\usepackage{times}
\usepackage{latexsym}
\usepackage{graphicx}

\usepackage{microtype}
\usepackage{verbatim}
\usepackage{array}

\newcolumntype{C}[1]{>{\centering\arraybackslash}p{#1}}

\aclfinalcopy % Uncomment this line for the final submission
\def\aclpaperid{1532} % Enter the acl Paper ID here

\title{Attention-based Contextual Language Model Adaptation for Speech Recognition }

\author[]{Richard Diehl Martinez}
\author[]{Scott Novotney}
\author[]{Ivan Bulyko} 
\author[]{\\Ariya Rastrow} 
\author[]{Andreas Stolcke}
\author[]{Ankur Gandhe}

\affil[]{Amazon Alexa, Seattle, WA, USA \authorcr \{\normalsize \tt mrtimri,snovotne,ibbulyko,arastrow,stolcke,aggandhe\}@amazon.com}

\date{}

\begin{document}
\maketitle
\begin{abstract}
Language modeling (LM) for automatic speech recognition (ASR) does not usually incorporate utterance level contextual information. For some domains like voice assistants, however, additional context, such as time at which an utterance was spoken, provides a rich input signal. We introduce an attention mechanism for training neural speech recognition language models on both text and non-linguistic contextual data \footnote{We make a large portion of our code available under: https://github.com/amazon-research/contextual-attention-nlm}. When applied to a large de-identified dataset of utterances collected by a popular voice assistant platform, our method reduces perplexity by 7.0\% relative over a standard LM that does not incorporate contextual information. When evaluated on utterances extracted from the long tail of the dataset, our method improves perplexity by 9.0\% relative over a standard LM and by over 2.8\% relative when compared to a state-of-the-art model for contextual LM.
\end{abstract}

\section{Introduction}

Conventional automatic speech recognition (ASR) systems include a language model (LM) and an acoustic model (AM). The LM component is trained separately, typically on large amounts of transcribed utterances that have been collected by an existing speech recognition system. 

Voice assistants have become ubiquitous and crucially rely on ASR systems to convert user inputs to text. They often collect utterances spoken by users, along with associated de-identified contextual information. We hypothesize that these additional data, such as the time at which an utterance was spoken, provide a useful input signal for a LM. As an example, knowing that an utterance was spoken on December 25th, a LM should learn that the word ``christmas" rather than ``easter" is more likely to follow the phrase ``lookup cookie recipes for".

To-date, some voice assistants have leveraged coarse geographic information for improving location search queries \cite{5495026, lloyd2012acoustic}. These past efforts, however, have largely focused on improving a particular skill of an ASR system, and not the system's speech recognition in general. 

In this paper, we focus on adapting recurrent neural network language models (RNN-LMs) to use both text and non-linguistic contextual data for speech recognition in general. While, outside of ASR, transformer-based \cite{vaswani2017attention} language models have largely replaced RNN-LMs, RNNs remain dominant in ASR architectures such as connectionist temporal classification \cite{graves2006connectionist}, and RNN-T \cite{graves2012sequence, he2019streaming}. 

The most common method for incorporating non-linguistic information into a RNN-LM is to learn a representation of the context that is concatenated with word embeddings as input to the model. This concatenation-based approach has been used in a variety of domains including text classification \cite{yogatama2017generative}, personalized conversational agents \cite{wen2013recurrent}, and voice search queries \cite{ma2018modeling}.
 
Recently, attention mechanisms, initially developed for machine translation \cite{bahdanau2014neural, luong2015effective}, have been used by neural LMs to adaptively condition their predictions on certain non-linguistic contexts. \citet{tang2016aspect} use an attention module that attends to word-location information to predict the polarity of a sentence. Similarly, \citet{zheng2019personalized} use an attention mechanism over learned personality trait embeddings, in order to generate personalized dialogue responses. 
 
The aforementioned approaches learn a representation of the context that is directly used as input to a neural LM. In contrast to these methods, \citet{jaech2018low} adapt the weight-matrix used in an RNN model to a given contextual input. The authors propose decomposing the weight-matrix into a set of left and right basis tensors which are then multiplied by a learned context embedding to produce a new weight-matrix. Factorizing the weight-matrix enables a larger fraction of a model's parameters to adjust to a given contextual signal. This factorization-based approach has proven effective in generating automatic completions of sentences that are personalized for particular users \cite{jaech2018personalized}.

%Outside of the domain of ASR, extensive research has been conducted on training neural language models using both text and non-linguistic contextual data. In this paper, we focus on adaptations to recurrent neural networks (RNNs), given their continued use in popular ASR models such as connectionist temporal classification \cite{graves2006connectionist}, and RNN-T \cite{graves2012sequence, he2019streaming}. 
%Building on the work of \citet{mikolov2012context} and \citet{jaech2018low}, -- work on transition 
We introduce an attention mechanism that augments both the concatenation-based and factorization-based approaches to condition a neural LM on context. The attention mechanism that we propose builds up a dynamic context representation over the course of processing an utterance. The resulting embedding can be used as an additional input to either the concatenation-based or factorization-based model. 

Our experiments focus primarily on conditioning neural LMs on datetime context. We concentrate on datetime information because of its widespread availability in many ASR systems. Our approach, however, can generalized to any type of context. To underscore this point we also provide results for conditioning LMs on geolocation information and dialogue prompts that are commonly available in ASR systems.

We evaluate our method on a large de-identified dataset of transcribed utterances. Compared to a standard model that does not include contextual information, using our method to contextualize a neural LM on datetime information achieves a relative reduction in perplexity of 7.0\%, and a relative reduction in perplexity of 9.0\% when evaluated on the tail of this dataset. Moreover, our attention mechanism can improve state-of-the-art methods for conditional LMs by over 2.8\% relative in terms of perplexity. %We find similar improvements when conditioning a neural LM on geolocation information and dialogue prompts.

\section{Data}

We use a corpus of over 5,000 hours of de-identified, transcribed English utterances, collected over several years. Each utterance also contains associated information about the year, month, day, and hour that the utterance was spoken. The datetime information is reported according to the local time zone of each given user. Any information about the device or the speaker from which an utterance originates has been removed. We randomly split our dataset into a training set, development set and test set, using a partition ratio of 90/5/5 and we ensure that each partition contains more than 500 hours worth of data.

%On average, an utterance in our dataset is 4.0 words long.

%8.7 million utterances, 260,000 utterances and 170,000 utterances, respectively. 

\section{Context Representation}

A typical utterance in our dataset might look like this: \\ \\
\centerline{2020-12-23 07:00 play christmas music.} \\ \\
In the example above, we can infer that the utterance ``play christmas music" was spoken on December 23, 2020 at 7 in the morning local time. In order to condition a LM on this datetime information, we consider two methods for transforming the contextual information into a continuous vector representation:
\begin{enumerate}
\item \textbf{Learned embeddings:} We first consider creating tokens for the month number, week number, day of the week and hour that an utterance was spoken. In the example above, we would transform the datetime information into tokens representing: month-12, week-52, wednesday, 7am. These tokens are subsequently used as input to the model, where they are passed through an embedding layer to generate context embeddings. These embeddings are initialized as random vectors, and trained along with the rest of the model. We experiment with different ways of parsing the information, such as encoding weekday versus weekend, or morning versus evening, but find this information is largely entailed within our method for processing datetime information.
\item \textbf{Feature-engineered representation:} Additionally, we consider transforming the datetime information into a single 8-dimensional feature-engineered vector, where the dimensions of the vector are defined as
$$\left[ \matrix{ sin(\frac{2\pi \cdot hour}{24}) & cos(\frac{2\pi \cdot hour}{24}) \cr \cr
sin(\frac{2\pi \cdot day}{7}) & cos(\frac{2\pi \cdot day}{7}) \cr \cr
sin(\frac{2\pi \cdot week}{53}) & cos(\frac{2\pi \cdot week}{53}) \cr \cr
sin(\frac{2\pi \cdot month}{12}) & cos(\frac{2\pi \cdot month}{12}) \cr} \right]$$
Since the datetime context is continuous and cyclical, this approach explicitly encodes temporal proximity in the date and time information.

\end{enumerate}

\subsection{Input Representation}
We assume as input to a model a sequence of either word or subword tokens, $w_{i}$ for $i \in \{1, \dots, n\}$, that are converted by an embedding layer into embeddings $x_{i} \in \mathcal{R}^{e}$ for $i \in \{1, \dots, n\}$, where \emph{n} is the length of the input sequence and \emph{e} is the dimensionality of the word embeddings. 

We additionally represent the contextual information as either:
\begin{enumerate}
 \item A set, $M$, of four learned context embeddings $M = \{m_1, m_2, m_3, m_4\}$, where $m_1$ is an encoding of the month information, $m_2$ is an encoding of the week information, $m_3$ is an encoding of the day of the week information, and $m_4$ is an encoding of the hour of the day information. When using the concatenation-based or factorization-based approaches without attention, we first concatenate the embeddings together, $m=[m_1;m_2;m_3;m_4]$, and use the resulting vector as input to the model.
\item A set, $M$, containing a single embedding $M = \{m\}$, where $m$ represents an 8-dimensional feature-engineered contextual datetime representation, as described in the previous section.
\end{enumerate}

\section{Model}

In this section, we first describe the architecture of the concatenation-based and factorization-based approaches. We then introduce our attention-mechanism that can be used to augment both of these approaches. The notation we use to describe architectures assumes a 1-layer RNN model. The methods we discuss, however, can be applied to each layer of a multi-layer RNN model.

\subsection{Concatenation-based LM Adaptation}

The concatenation-based approach learns a weight matrix $\mathbf{W_{m}}$ of dimensionality $\mathcal{R}^{f \times d}$, where \emph{f} represents the size of the context representation and \emph{d} represents the hidden-dimensionality of the RNN model. In practice, \emph{f} is either a hyperparameter when datetime context is represented as learned embeddings, or $f = 8$ when this context is represented as a feature-engineered vector. When representing contextual information as learned embeddings, recall that we first concatenate the embeddings together before passing these into the model. In this case, \emph{f} is four-times the size of each individual context embedding.

A standard RNN model without contextual information keeps track of a hidden-state at time-step \emph{t}, $h_t$, that is calculated as
$$ 
h_t = \sigma \big( \mathbf{W_x} x_t + \mathbf{W_h} h_{t-1} + b \big),
$$
where $x_t$ represents the word embedding at time-step \emph{t}, \emph{b} is a bias vector, $\mathbf{W_x} \in \mathcal{R}^{e \times d}$, and $\mathbf{W_h} \in \mathcal{R}^{d \times d}$.

In the concatenation-based approach, this hidden-state is adapted in the following manner 
$$ 
h_t = \sigma \big( \mathbf{W_{m}} m + \mathbf{W_x} x_t + \mathbf{W_h} h_{t-1} + b \big).
$$
Notice that the expression above can be equivalently calculated by concatenating the matrices $\mathbf{W_{m}}$ and $\mathbf{W_x}$, as well as the vectors $m$ and $x_t$ 
$$ 
h_t = \sigma \big( [\mathbf{W_x};\mathbf{W_{m}}] [x_t; m] + \mathbf{W_h} h_{t-1} + b \big).
$$
To generate a prediction, $\hat{y}_t$ for a word at time-step \emph{t}, $h_t$ is passed through a projection layer, $W_v \in \mathcal{R}^{d \times |V|}$ to match the dimension of the vocabulary size $|V|$, before applying a softmax layer
$$\hat{y}_t = softmax \big( W_v h_t\big).$$

\subsection{Factorization-based LM Adaptation}

Unlike the concatenation-based approach, which directly inserts contextual information into the RNN cell, the factorization-based method adapts the weight matrices $\mathbf{W_x}, \mathbf{W_h}$ of the RNN model. Compared to the concatenation-based architecture, this approach adapts a larger fraction of the RNN model's parameters.

The adaption process involves learning basis tensors $\mathbf{W^{(L)}_{x^\prime}}, \mathbf{W^{(R)}_{x^\prime}}$ and $\mathbf{W^{(L)}_{h^\prime}}, \mathbf{W^{(R)}_{h^\prime}}$. These basis tensors are of fixed rank \emph{r}, where \emph{r} is a tuned hyperparameter. The left adaptation tensors, $\mathbf{W^{(L)}_{x^\prime}}, \mathbf{W^{(L)}_{h^\prime}}$, are of dimensionality $\mathcal{R}^{f \times e \times r }$, and $\mathcal{R}^{f \times d \times r }$, respectively. The right adaptation tensors, $\mathbf{W^{(R)}_{x^\prime}}, \mathbf{W^{(R)}_{h^\prime}}$ are both of dimensionality $\mathcal{R}^{r \times d \times f}$. We can now use the contextual representation to interpolate the two sets of basis tensors to produce two new weight matrices, $\mathbf{W^\prime_x}$ and $\mathbf{W^\prime_h}$, where
 
$$\mathbf{W^\prime_x} = \mathbf{W_x} + (\mathbf{W^{(L)}_{x^\prime}}^Tm)^T(\mathbf{W^{(R)}_{x^\prime}}^Tm)$$
$$\mathbf{W^\prime_h} = \mathbf{W_h} + (\mathbf{W^{(L)}_{h^\prime}}^Tm)^T(\mathbf{W^{(R)}_{h^\prime}}^Tm).$$

The resulting matrices $\mathbf{W^\prime_x}, \mathbf{W^\prime_h}$ are now used as the weights in the RNN cell. A prediction, $\hat{y}_t$, is generated in the same manner as in the concatenation-based model.

\subsection{Attention Mechanism}

We propose an attention mechanism that augments both the concatenation-based and factorization-based approaches. We apply this mechanism to the context embeddings at each time-step of the RNN model, in order to adapt the context representation dynamically. We hypothesize that at certain time-steps within an utterance, attending to particular datetime information will facilitate the model's predictions more than other information.

For instance, assume a LM is given the phrase ``what temperature will it be on friday". By the time the model has observed the words ``what temperature will", we would expect the model to condition the predictions of the remaining words primarily on the hour and day information. Using an attention mechanism enables us to dynamically weight the importance that the model places on particular datetime context as the model processes an utterance.

We assume as input to the attention mechanism the same set $M$ of context representations. However, in the case where datetime information is represented as a feature-engineered vector, we augment $M$ to include an 8-dimensional vector of all 0s: $M = \{m, \textbf{0}\}$. We do so because our attention mechanism builds a dynamic representation of the context by interpolating over multiple context embeddings. Thus, the attention mechanism can act as a learnable gate to limit the non-linguistic context passed into the model. We also experiment with adding a similar vector of all 0s in the case where context embeddings are learned, but find no improvement.

\begin{figure}[h!]
 \centering
 \includegraphics[scale=0.22]{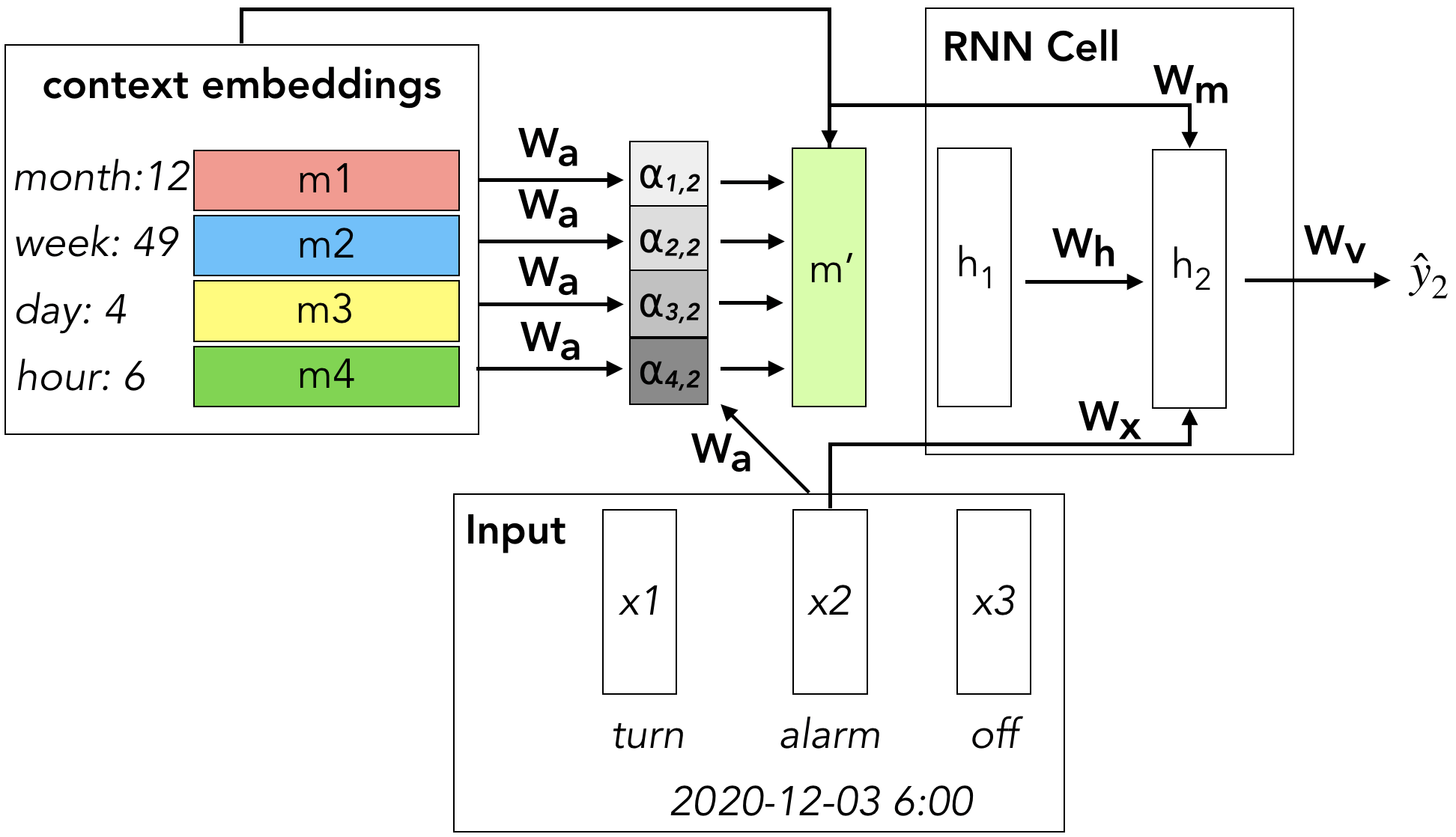}
 \caption{Model architecture of the concatenation-based model using attention. Datetime context is encoded as learned embeddings, and the input word embedding at time-step \emph{t} is used as the query vector at \emph{t}.}
 \label{fig:model_architecture}
\end{figure}

In addition to the set $M$, the attention mechanism takes in a query vector, $q_t$, for each time-step \emph{t}. We propose two methods for defining this query vector
\begin{enumerate}
	\item Let $q_t = x_t$, where $x_t$ is the embedding for the input word at time-step \emph{t}. 
	\item Let $q_t = h_t$, where $h_t$ is the hidden state of the RNN model at time-step \emph{t}.
\end{enumerate}

\begin{table*}
\centering
\begin{tabular}{ |p{1.5cm}|p{3cm}|p{2.3cm}||p{1.2cm}|p{1.2cm}|p{1.2cm}|}
\hline
\center{Method} & \center{Context Approach} & \center{Attention Query} & \multicolumn{3}{c|}{Relative Perplexity Reduction} \\
\cline{4-6}
& & & Full & Head & Tail \\
\hline \hline 
Default & NA & NA & 0\% & 0\% & 0\% \\
\hline
Prepend & Embeddings & NA & 0.83\% & \textbf{2.54\%} & 0.12\% \\
Prepend & Feature-engineered & NA & \textbf{1.20\%} & 2.30\% & \textbf{0.74\%} \\
\hline
Concat & Embeddings & NA & 6.82\% & 2.82\% & 8.68\% \\
Concat & Embeddings & Hidden & 7.00\% & 2.83\% & 8.91\% \\
Concat & Embeddings & Word & \underline{\textbf{7.02\%}} & \underline{\textbf{2.89\%}} & \underline{\textbf{8.96\%}} \\
Concat & Feature-engineered & NA & 6.82\% & 2.65\% & 8.71\% \\
Concat & Feature-engineered & Hidden & 7.0\% & 2.71\% & 8.97\% \\
Concat & Feature-engineered & Word & 6.94\% & 2.60\% & 8.94\% \\
 \hline
Factor & Embeddings & NA & 3.29\% & 2.26\% & 3.93\% \\
Factor & Embeddings & Hidden & 4.82\% & 2.53\% & 5.96\% \\
Factor & Embeddings & Word & 5.40\% & \textbf{2.58\%} & 6.71\% \\
Factor & Feature-engineered & NA & 5.44\% & 2.00\% & \textbf{7.10\%} \\
Factor & Feature-engineered & Hidden & \textbf{5.57\%} & 2.31\% & 6.82\% \\
Factor & Feature-engineered & Word & 5.05\% & 2.25\% & 6.31\% \\
\hline
\end{tabular}
\caption{Test set perplexities of contextual LMs based on datetime information, with relative reductions compared to the \emph{Default} LSTM model that does not use contextual datetime information. We bold best results within each type of method, and underline best results overall. Improvements in perplexity from using our attention mechanism are statistically significant.}
\label{tab:ppl}
\end{table*}

Importantly, when $q_t$ is chosen such that $q_t = x_t$, we can parallelize the computation of the attention mechanism for all time-steps before running a forward pass through the model. This cannot be done when $q_t = h_t$, as the attention mechanism can only be computed sequentially for each hidden state of the model.

Regardless of the choice of $q_t$, the attention mechanism first computes a score for each context embedding $m_i \in M$ for a given $q_t$. To compute this score, we learn a weight matrix $W_a$. The size of $W_a$ is $\mathcal{R}^{f \times e}$ if $q_t = x_t$, or $\mathcal{R}^{f \times d}$ if $q_t = h_t$. 

For a given $m_i \in M$ and $q_t$, we calculate a score as $$score(m_i, q_t) = m_i^T\mathbf{W_a}q_t.$$
We then define the alignment score as
$$\alpha_{i, t} = softmax(score(m_i, q_t))$$
The alignment scores are finally used to build up a dynamic representation of the context, $m_t^\prime$, for a given time-step. 
 $$m_t^\prime = \sum_{i=1}^{|M|} \alpha_{i, t} m_i$$ 
 
 We can now use this constructed context, as the context input to either the concatenation-based or the factorization-based approach. 

In Figure \ref{fig:model_architecture} we illustrate how the attention mechanism augments the concatenation-based approach. For an utterance like ``turn alarm off", we showcase how the model builds a dynamic representation of the datetime context, at a given time-step \emph{t} ( \emph{t} = 2 in the figure). 

\section{Experimental Setup}
We used a 1-recurrent-layer LSTM model \cite{hochreiter1997long} as the base model in all of our experiments. Both the concatenation-based and factorization-based methods can be easily adapted to use an LSTM cell. Models were trained using the Adam optimizer with an initial learning rate of 0.001, and a standard cross entropy loss function. Each of the LMs was trained for 400,000 batch update steps, using a batch-size of 256. The training of each model was conducted on a single V100 GPU, with 16GB of memory on a Linux cluster, and took roughly 6 hours to train. Implementation of the model and training procedure was written in PyTorch and native PyTorch libraries.
We used a fixed dimensionality of 512 for word, context and hidden state embeddings. We initially experimented with smaller and larger embedding sizes (50, 100, 1024), but found that 512 generally provided a good tradeoff between model performance and compute resources required to train a model. We set the rank of the basis tensors in the factorization-approach to 5, after experimenting with rank sizes 2, 3, 10, 15, 20. In practice, we found that the larger the rank size the less stable the training procedure became. Other hyperparameters, such as the initial learning rate, were selected via random search. We initialized random weights using Xavier-He weight initialization \cite{he2015delving}.

\section{Results}
\subsection{Datetime}

%We evaluate our model on a heldout set of 300,000 utterances that are randomly sampled from the full set of collected utterances. The utterances in our train and evaluation set were collected in the same time-range from 2013-2019. We additionally partition the development set into a head and tail set, representing respectively the top 5\% most frequently occurring utterance and utterance that only occur once. The head and tail sets each encompass around 100,000 utterances. 

We evaluated our models on a heldout set of utterances that were randomly sampled from the full dataset. The utterances in our training and evaluation set were collected in the same time-range. We also defined the head and tail subsets of our development set, representing, respectively, the top 5\% most frequently occurring utterances, and utterances occurring only once.

\begin{figure*}[h!]
 \centering
 \includegraphics[scale=0.65]{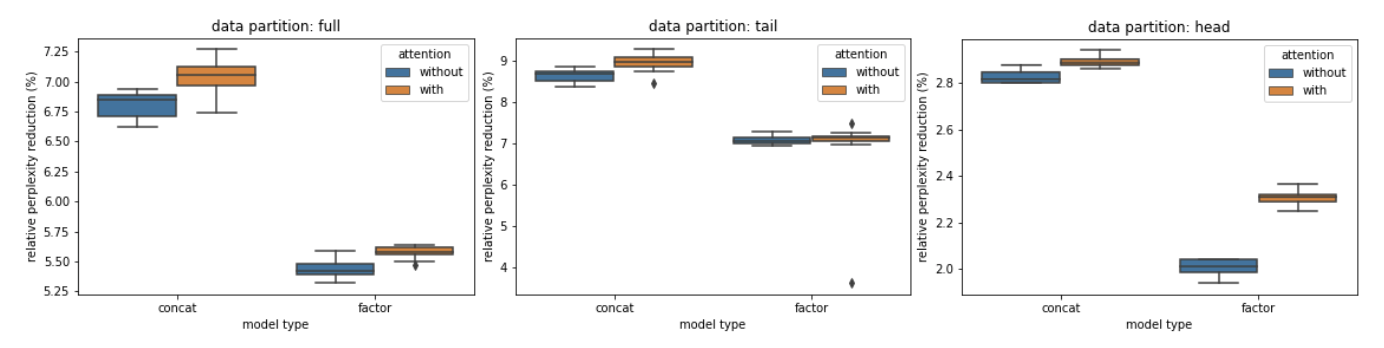}
 \caption{Perplexity confidence bounds at a 95\% confidence level for the best-performing concatenation-based and factorization-based models with and without attention. Bounds are evaluated on the full, head and tail partitions of the evaluation set. Perplexity reductions are relative to the \emph{Default} LSTM model that does not use contextual datetime information.} 
 \label{fig:ci}
\end{figure*}

We used two metrics for our evaluations: perplexity and word error rate. Perplexity is a common statistic widely used in language modeling and speech recognition to measure how well a language model predicts a sample of text \cite{jelinek1977perplexity}. Word error rate, on the other hand, measures the Levenshtein (minimum edit) distance between a recognized word sequence and a reference word sequence. In practice, these two statistics have been shown to be correlated by a power law relationship \cite{klakow2002testing}.

In Table \ref{tab:ppl}, we report the relative decrease in perplexity of models that leveraged datetime context compared to a baseline LSTM model that did not use any contextual information. We additionally trained a simple baseline, \emph{Prepend}, which was comprised of a standard LSTM model that treated datetime context as input tokens that were prepended to the input texts. 

In reporting our results, we distinguish between the two forms of representing contextual information: either as learned embeddings or as a feature-engineered representation. We also differentiate between the two variants of encoding the query vector used by the attention mechanism: either by using the hidden state vector, or the input embedding at a given time-step.

For the concatenation-based model, we found that adding our attention mechanism led to further reductions in perplexity, regardless of the type of query vector or context representation used. We obtained the best results when representing datetime information as learned embeddings and using the input embedding at a given time-step as the query vector.

%\begin{table*}
%\centering
%\begin{tabular}{ |c|c|c|c|c|c|}
%\hline
%Method & Metadata Approach & Attention Context & \multicolumn{3}{c|}{Confidence Interval} \\
%\cline{4-6}
%& & & Full & Head & Tail \\
%\hline \hline
%Concat & Embeddings & NA & (17.41, 17.44) & (7.56, 7.57) & (25.25, 25.31)\\
%Concat & Embeddings & Word & (17.36, 17.40) & (7.55, 7.56) & (25.17, 25.24) \\
%\hline \hline
%Factor & Feature-engineered & NA & (17.67, 17.69) & (7.62, 7.63) & (25.70, 25.74)\\
%Factor & Feature-engineered & Hidden & (17.65, 17.66) & (7.60, 7.61) & (25.60, 25.99) \\
%\hline
%\end{tabular}
%\caption{We report lower and upper confidence bounds at a 95\% confidence level for the best performing concatenation-based and factorization-based models with and without attention. Confidence bounds first specify the lower bound, then the upper bound.} 
%\label{tab:ci}
%\end{figure*}

We corroborated these results by computing 95\% confidence intervals for the best-performing concatenation-based models with and without attention. Confidence intervals were calculated by running the training algorithm 10 times for each model type. Figure \ref{fig:ci} visualizes the intervals. 
 
In the case of the factorization-based approach, we achieved the lowest perplexity when the attention mechanism used the hidden state of the RNN model as the query vector and datetime information was represented as a feature-engineered vector. Again, we found that on the full dataset the improvement in perplexity by using our attention mechanism was statistically significant. 

In nearly every experiment we ran, we found that our attention mechanism further reduced perplexity. The use of attention led to the largest relative improvement in the factorization-based approach when using learned context embeddings. In this instance, perplexity was reduced by 2.8\% on the tail of our evaluation set, and by 2.1\% on the full dataset.

\begin{table}[h!]
\centering
\begin{tabular}{ |c|c|c|p{0.7cm}|p{0.7cm}|}
\hline
Method & Context & Attention & \multicolumn{2}{c|}{WERR (\%)} \\
\cline{4-5}
& Approach & Query & Full & Tail \\
\hline \hline
Default & NA & NA & 0.0 & 0.0 \\
Prepend & FE & NA & 0.0 & 0.0 \\
Concat & Emb & Word & \textbf{1.1} & \textbf{1.2} \\
Factor & FE & Hidden & 0.8 & 0.8 \\
\hline
\end{tabular}
\caption{We report relative improvement (decrease) in WER compared to the \emph{Default} LSTM model that does not use contextual information. Context representation approaches are feature-engineered (FE) or embeddings (Emb).} 
\label{tab:wer}
\end{table}

In addition to evaluating the models on relative reductions in perplexity, we also validated the downstream performance of a hybrid CTC-HMM \cite{graves2006connectionist} ASR system that incorporated contextual information in its LM component. As the LM component of this system, we used the best-performing models within each category of method that we report in Table \ref{tab:ppl}. Table \ref{tab:wer} summarizes the results. We evaluated the relative WER reduction(WERR) on a large test set of de-identified, transcribed utterances representative of general user interactions with Alexa, as well as on the tail of this dataset. As in Table \ref{tab:ppl}, the concatenation-based model with attention mechanism achieved the largest reductions in WER.

\begin{comment}
\begin{table*}[ht!]
\centering
\begin{tabular}{ |c|c|c|c|c|c|c|}
\hline
Context Type & \multicolumn{3}{c|}{Relative Perplexity (\%)} & \multicolumn{3}{c|}{WERR (\%)} \\
\cline{2-7}
& Full & Head & Tail & Full & Head & Tail \\
\hline \hline
Default & 0.0 & 0.0 & 0.0 & 0.0 & 0.0 & 0.0 \\
Datetime & 11.4 & 8.7 & 11.6 & 1.6 & \underline{2.1} & 1.7 \\
Geo-hash & 12.4 & \underline{11.6} & 12.5 & 0.5 & -1.4 & 1.0 \\
Dialogue Prompt & \underline{13.9}	 & 11.3 & \underline{14.1} & 0.3 & -2.8 & 0.6 \\
All & 7.3 & 4.8 & 7.5 & \underline{2.1} & 0.0 & \underline{2.6} \\
\hline
\end{tabular}
\caption{For each type of contextual information, we report relative improvement (decrease) in WER and (decrease) in perplexity compared to the \emph{Default} LSTM model that does not use contextual information. We report results on both the full test dataset as well as utterances from the tail of the dataset. The final row shows the result of combining together the datetime, geo-hash and dialogue prompt contexts. The best results are underlined.} 
\label{tab:other-context}
\end{table*}
\end{comment}

\begin{table}[ht!]
\centering
\begin{tabular}{ |c|c|c|c|c|}
\hline
Context Type & \multicolumn{2}{C{2cm}|}{Relative PPL Reduction (\%)} & \multicolumn{2}{c|}{WERR (\%)} \\
\cline{2-5}
& Full & Tail & Full & Tail \\
\hline \hline
Default & 0.0 & 0.0 & 0.0 & 0.0 \\
Datetime & 11.4 & 11.6 & \underline{1.6} & \underline{1.7} \\
Geo-hash & 12.4 & 12.5 & 0.5 & 1.0 \\
Dialogue Prompt & \underline{13.9}	 & \underline{14.1} & 0.3 & 0.6 \\
\hline
\end{tabular}
\caption{We report relative improvement (decrease) in WER and (decrease) in perplexity (PPL) compared to the \emph{Default} LSTM model that does not use contextual information. We report results on both the full test dataset as well as utterances from the tail of the dataset. The best results are underlined.} 
\label{tab:other-context}
\end{table}

\subsection{Other Non-Linguistic Context}

So far, our experiments have focused exclusively on conditioning neural LMs on datetime context. We underscore, however, that the contextual mechanism we introduce can be applied to any type of contextual information that can be represented as embeddings. To illustrate this point we train two neural LMs using two other types of context: geolocation information and dialogue prompts.

 We train the LMs on a subset of the utterances of the initial dataset which also contain utterance-level geo-hash information and dialogue prompt information. The geo-hash information \footnote{We use a two integer precision geo-hash.} associated with each utterance encodes a very rough estimate of the geolocation of a user's device. Dialogue prompts indicate whether a transcribed utterance was an initial query to the dialog system or if it was a follow-up turn.

%For instance, all devices in the US state of Massachusetts share the same geo-hash. 
We learn embeddings to represent both the geo-hash and the dialogue prompt information. We ingest both types of contexts via the concatenation-based approach, using word-embeddings as the attention queries. We evaluate these models on a test set of de-identified utterances representative of user interactions with Alexa. Table \ref{tab:other-context} summarizes the results. In general, we find that conditioning neural LMs on each of the different types of context reduces perplexity and WER.

%Finally, we also experiment with concatenating together datetime, geo-hash and dialogue prompt embeddings, which are each generated by a separate attention mechanism, before ingesting the combined embedding into the neural LM. 

\section{Analysis}

In this section, we focus once again on datetime information to better understand how contextual LMs use datetime signal.

\subsection{Datetime Context Signal}

The first question we hope to answer is: to what extent can the relative improvements in perplexity and WER in the models that incorporate datetime context be explained by the additional signal from the context versus the additional parameters that these models contain?

To answer this question, we randomly shuffled the datetime information associated with each utterance in our training and test sets. For each of our best-performing models in a given category of method (prepend, concat, or factor), we retrained and evaluated those models on the dataset containing shuffled datetime information.

In Table \ref{tab:random}, we report the relative degradation (i.e., a negative reduction) in perplexity resulting from evaluating these models on the shuffled datetime contexts. In general, if a model uses datetime information as an additional signal, we would expect the performance of the model to decrease when the datetime context is shuffled.

\begin{comment} 
\begin{table}[h!]
\centering
\begin{tabular}{ |p{1cm}|p{1.26cm}|p{1.2cm}|p{0.5cm}|p{0.6cm}|p{0.1cm}|}
\hline
Method & Context & Attention & \multicolumn{3}{c|}{Rel. PPL Inc. (\%) } \\
\cline{4-6}
& Approach & Query & Full & Head & Tail\\
\hline \hline
Prepend & FE & NA & 1.5 & 2.4 & 1.2 \\
Concat & Emb & Word & 1.6 & 2.7 & 1.2 \\
Factor & FE & Hidden & 1.4 & 2.3 & 1.0 \\
\hline
\end{tabular}
\caption{Relative degradation (increase) in perplexity (PPL) of models that incorporate datetime information when that context is randomly shuffled. Context representations are feature-engineered (FE) or embeddings (Emb).} 
\label{tab:random}
\end{table}
\end{comment}

\begin{table}[h!]
\centering
\begin{tabular}{ |c|C{1.51cm}|C{1.31cm}|c|c|}
\hline
Method & Context Approach & Attention Query & \multicolumn{2}{C{1.97cm}|}{Relative PPL Reduction (\%) } \\
\cline{4-5}
& & & Full & Tail\\
\hline \hline
Prepend & FE & NA & -1.5 & -1.2 \\
Concat & Emb & Word & -1.6 & -1.2 \\
Factor & FE & Hidden & -1.4 & -1.0 \\
\hline
\end{tabular}
\caption{Relative degradation in perplexity (PPL) of models that incorporate datetime information when that context is randomly shuffled. Context representations are feature-engineered (FE) or embeddings (Emb).} 
\label{tab:random}
\end{table}

We observed the overall largest relative degradation in perplexity, when using the concatenation-based model. Recall that when trained on correct datetime information this was our best-performing model overall in terms of both perplexity and WER, indicating that the performance of this model can be attributed in part to its use of contextual information. %However, regardless of model type, all models showed substantial degradation in performance when trained and evaluated on shuffled datetime contexts. 

%For these three models, we additionally compute which particular utterances decrease the most in perplexity, when compared to a model that does not use metadata information. Uniformly across model types, we find that included in the top 20 utterances which decrease most in perplexity are "snooze", "happy thanksgiving", "halloween" and "christmas". This lends further evidence that our contextual models have learned to condition their predictions on datetime information.

\subsubsection{Visual Analysis}

\begin{figure}[h!]
 \centering
 \includegraphics[scale=0.48]{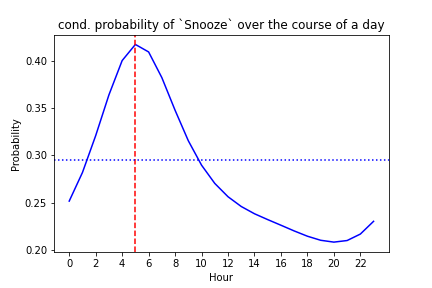}
 \caption{Changing conditional probability of the word ``snooze" as the associated hour of day information varies.}
 \label{fig:lineplot}
\end{figure}

In addition to these results, we visualize how the contextual LMs leverage datetime contexts. For a given utterance, we can evaluate the probability of the words in the utterance as we vary the datetime information associated with the utterance. In Figure \ref{fig:lineplot}, we evaluate the conditional probability of the word ``snooze" in an utterance following the start-of-sentence token, as we vary the hour of day information associated with this utterance. As we would expect, the probability of this ``snooze" is highest in the morning (between 5 and 6 am), as users are waking up and snoozing their alarms. As we move away from the morning hours, the conditional probability of the word ``snooze" decreases substantially, reaching a low-point by the afternoon and evening. The horizontal blue dashed line indicates the conditional probability of the word 'snooze' following the start-of-sentence token when evaluated with a LM that does not ingest datetime information. This analysis further corroborates that the trained contextual LMs successfully condition their predictions on datetime information.

\subsection{Attention Weights}

\begin{figure}[h!]
 \centering
 \includegraphics[scale=0.55]{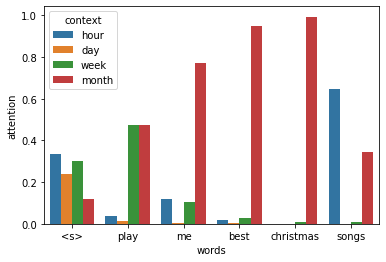}
 \caption{Changing attention weights placed on particular context embeddings over the course of an utterance.}
 \label{fig:barplot}
\end{figure}

We next seek to understand how the attention mechanism constructs a dynamic representations of datetime context. To do so, we visualize the weights of the attention mechanism as an utterance is processed by the model. For a given utterance like ``play me best christmas songs" spoken in December, we highlight the changing weight placed on each of the datetime information. Figure \ref{fig:barplot} shows this analysis.

When the model processes the start-of-sentence token, the attention mechanism weights each of the datetime information roughly equally. However as the model processes the subsequent words ``play me best", the attention begins to shift towards using more of the month information (i.e., that this utterance was spoken in December), and away from hour and day information. This would suggest that conditioning on the fact that the utterance was spoken in December can help the model predict the type of media to play. 

Once the model observes the word ``christmas", it places all of the attention on the month information, indicating the model has successfully learned that ``christmas" is a word strongly associated with a particular month (i.e., December). Finally when the word ``songs" is ingested, the model substantially reduces the weight placed on month information and in turn increases the weight on hour information. This shift might indicate that the model has learned to condition the type of music users listen to to the hour of the day. Overall, the behavior of the attention mechanism is consistent with our initial hypothesis that certain types of datetime information can benefit a contextual LM model more than others over the course of an utterance. 

\begin{comment}
In Figure \ref{fig:barplotrad} we repeat this analysis, using the feature-engineered datetime representation, and the hidden state of the RNN model as the query vector. Recall that when using this method for encoding datetime information, we additionally incorporate a null vector of 0s that the attention model can attend to, in order to ignore contextual information passed into the model. Over the course of an utterance, the model can attend to this vector when contextual information is not useful for predicting the next word in the utterance. As input to the model, we pass in the utterance ``what temperature will be on friday". 

As in the previous example, when the model processes the start-of-sentence token, the attention mechanism places equal weight on the datetime information as on the null embedding. However, once the models reads in the first couple of words ``what temperature", the attention mechanism attends nearly exclusively to the datetime context. We can infer that the model learns to associate temperature-related queries with a particular time or day. This analysis is further reinforced when the model again places a large fraction of the attention weight on the datetime context after processing the word ``on" and before predicting ``friday". 

\begin{figure}[h!]
 \centering
 \includegraphics[scale=0.55]{barplot_radians.png}
 \caption{Changing attention weights placed on a feature-engineered context vector over the course of an utterance.}
 \label{fig:barplotrad}
\end{figure}

\end{comment}

\section{Related Work}

Within the domain of ASR, \citet{biadsy2017effectively} have explored using an adaptive-training approach to incorporate non-linguistic features into a maximum entropy LM. They propose first training the parameters of a LM that are associated with text data, then freezing those parameter and learning parameters associated with multiple types of non-linguistic features. 

\citet{zhang2019neural} and \citet{yoon2017efficient} propose a similar two-pronged approach for personalizing conversational neural LMs. They propose first pretraining a RNN-LM on a large dataset of conversational data, then finetuning the model on data associated with a particular user.

More recently, \citet{jain2020contextual} and \citet{liu2020contextualizing} propose attention mechanisms for conditioning RNN-T and hybrid ASR systems on words that are likely to occur in an utterance.

Another related line of research has explored learning utterance embeddings for dialogue systems using Gaussian mixture models that are enhanced with utterance-level context, such as intent \cite{yan2020unknown}. 

Outside of ASR, our work directly builds upon the concatenation-based \cite{mikolov2012context} and factorization-based \cite{jaech2018low} approaches to condition RNN-LMs on sentence context. The concatenation-based approach has been adopted as a common method for incorporating non-linguistic context into a neural LM \cite{yogatama2017generative, wen2013recurrent, ma2018modeling, ghosh2016contextual}. Methods that apply low-rank matrix factorization to RNNs are somewhat newer, and were first explored by \citet{kuchaiev2017factorization}.

Our contribution lies first in the application of these models to ASR, and secondly their extension with an attention mechanism. The attention mechanism we propose builds on the global attention model proposed by \citet{luong2015effective}. While attention-based models have been used to condition neural models on particular aspects or traits \cite{zheng2019personalized, tang2016aspect}, we focus on contextual information that benefits ASR systems. %These models have tended to use a Bahdanau attention mechanism \cite{bahdanau2014neural}. In practice we find the simpler, general global attention mechanism to yield strong results.

\section{Conclusion}

In this paper, we introduce an attention-based mechanism to condition neural LMs for ASR on non-linguistic contextual information. The proposed model dynamically builds up a representation of contextual information that can be ingested into a RNN-LM via a concatenation-based or factorization-based approach. We find that incorporating datetime context into a LM can yield a relative reduction in perplexity of 9.0\% over a model that does not incorporate context. Moreover, the attention mechanism we propose can improve state-of-the-art contextual LM models by over 2.8\% relative in terms of perplexity. 
While we focus on datetime information, we demonstrate that our approach can be applied to any type of non-linguistic context, such as geolocation and dialogue prompts.

\bibliographystyle{acl_natbib}
\bibliography{acl2021}

%\appendix

\end{document}

\maketitle

\section{Response to reproducibility checklist}

In our paper, we collect a dataset of user utterances on the Alexa voice assistant platform. Each of the utterances in our dataset contains associated contextual information, such as the date and time that the utterance was spoken. Any personally identifiable information about a user has been anonymized. Unfortunately, we are unable to release this dataset, in full or in part, due to customer privacy concerns. 

The contribution of our paper lies in the general-purpose methodology we introduce for incorporating non-linguistic contextual information into a language model. The methods we propose can be used for any type of dataset that contains associated contextual information. 

We also introduce a number of computational experiments to validate the claims we make in our paper. The `Model' section of our paper along with the `Experimental Setup' section provide full details to enable the reproduction of our results. The attention mechanism we introduce is fully specified in the `Model' section of our paper, along with the two contextual (concatenation-based and factorization-based) language models that we build upon. We are currently working on obtaining legal permission to open-source our code from our institution. Unfortunately, we are unlikely to obtain this permission by the deadline for submitting the technical appendix.